\documentclass[sigconf,svgnames]{acmart}

\usepackage{booktabs} 
\usepackage{threeparttable}
\usepackage{colortbl} 
\usepackage{color}
\usepackage{textcomp}
\usepackage{soul} 
\usepackage{amsmath}
\usepackage{comment}
\usepackage{multirow}
\usepackage{algorithm} 
\usepackage{algpseudocode} 

\usepackage{amsfonts}
\usepackage{bm}
\usepackage{stfloats}

\makeatletter
\g@addto@macro\normalsize{%
  \abovedisplayskip 2pt plus 1pt minus 2pt%
  \belowdisplayskip \abovedisplayskip
  \abovedisplayshortskip 2pt plus1pt  minus2pt%
  \belowdisplayshortskip 2pt plus1pt minus2pt%
}

\makeatother

\newtheorem{myDef}{Definition}

\setcopyright{acmcopyright}
\copyrightyear{2019} 
\acmYear{2019} 
\acmConference[CIKM '19]{The 28th ACM International Conference on Information and Knowledge Management}{November 3--7, 2019}{Beijing, China}
\acmBooktitle{The 28th ACM International Conference on Information and Knowledge Management (CIKM '19), November 3--7, 2019, Beijing, China}
\acmPrice{15.00}
\acmDOI{10.1145/3357384.3357989}
\acmISBN{978-1-4503-6976-3/19/11}

\begin{document}
\fancyhead{}

\title[Cross-domain Aspect Category Transfer and Detection]{Cross-domain Aspect Category Transfer and Detection via Traceable Heterogeneous Graph Representation Learning} 

\author{Zhuoren Jiang}
\authornote{These two authors contributed equally to this research.}
\affiliation{%
  \institution{School of Data and Computer Science}
  \institution{Sun Yat-sen University}
  \city{Guangzhou}
  \state{China}
}
\email{jiangzhr3@mail.sysu.edu.cn}

\author{Jian Wang}
\authornotemark[1]
\email{eric.wj@alibaba-inc.com}
\author{Lujun Zhao}
\email{lujun.zlj@alibaba-inc.com}
\author{Changlong Sun}
\email{changlong.scl@taobao.com}
\affiliation{%
  \institution{Alibaba Group}
  \city{Hangzhou}
  \state{China}
}

\author{Yao Lu}
\affiliation{%
  \institution{School of Data and Computer Science}
  \institution{Sun Yat-sen University}
  \city{Guangzhou}
  \state{China}
}
\email{luyao23@mail.sysu.edu.cn}

\author{Xiaozhong Liu}
\authornote{Corresponding author}
\affiliation{%
  \institution{School of Informatics, Computing and Engineering}
  \institution{Indiana University Bloomington}
  \city{Bloomington, IN}
  \country{USA}}
\email{liu237@indiana.edu}

\begin{abstract}
Aspect category detection is an essential task for sentiment analysis and opinion mining. However, the cost of categorical data labeling, e.g., label the review aspect information for a large number of product domains, can be inevitable but unaffordable. In this study, we propose a novel problem, cross-domain aspect category transfer and detection, which faces three challenges: various feature spaces, different data distributions, and diverse output spaces. To address these problems, we propose an innovative solution, Traceable Heterogeneous Graph Representation Learning (THGRL). Unlike prior text-based aspect detection works, THGRL explores latent domain aspect category connections via massive user behavior information on a heterogeneous graph. Moreover, an innovative latent variable ``Walker Tracer'' is introduced to characterize the global semantic/aspect dependencies and capture the informative vertexes on the random walk paths. By using THGRL, we project different domains' feature spaces into a common one, while allowing data distributions and output spaces stay differently. Experiment results show that the proposed method outperforms a series of state-of-the-art baseline models.
\end{abstract}

%
%



\keywords{Aspect Category Detection; Cross-domain Aspect Transfer; Heterogeneous Graph Representation Learning}

\maketitle

\section{Introduction}\label{sec:intro}
As an essential task of aspect-based sentiment analysis \cite{pontiki2016semeval}, aspect category detection provides important means to identify the refereed aspects from a free-text review \cite{zhou2015representation}. For instance, in a review of \textit{``The length of dress is perfect for me, it goes right above knee. At this price it is a bargain...''}, the aspect categories \textit{``price''} and \textit{``wearing effect''} can be detected for further sentiment analysis and opinion mining.

To address the aspect detection problem, a number of machine learning approaches \cite{ganu2009beyond,kiritchenko2014nrc,zhou2015representation,ruder2016insight}, i.e., learning a specific aspect classifier based on review representation for each target domain, have been proposed. Although the mathematical mechanism behind the models can be different, they all share the same prerequisite: a decent labeled training dataset is available for the target product domain. Unfortunately, such cost can be inevitably high for e-commerce platforms because different product domains may have different aspect categories and users may use distinctly different words to express the same aspect category across different domains. Take the world leading e-commerce services, eBay and Taobao, as an example. They both feature more than 20,000 product domains, and the cost of data annotation can be very high. 


Intuitively, facing a (target) domain with limited labeled training data (cold start problem), one could improve the model's performance by enhancing the training process with supplementary labeled data from a related (source) domain. However, in the e-commerce scenario, the following reasons make the cross-domain aspect knowledge transfer a challenging task: (1) \textbf{Various feature (vocabulary) spaces}. While shopping in different domains, customers may use different vocabularies to review the target products or services. (2) \textbf{Different data distributions}. Even for the same aspect, term usage can vary in different domains. For instance, for aspect ``size,'' customers will use phrases such as \textit{``extra large (XL)''} in clothing domain, while \textit{``270mm''} in shoes domain. (3) \textbf{Diverse output (aspect) spaces}. Each domain can have its own aspect categories, e.g., aspect \textit{``pilling''} for clothing, and aspect \textit{``functionality''} for shoes.  

Because of these reasons, most existing transfer learning methods can hardly be applied to address cross-domain aspect transfers. For instance, existing Domain Adaption (DA) models\cite{glorot2011domain,pan2010cross,zellinger2017central}---which train and test models on different distributions cross domains---can hardly be adopted for cross-domain aspect category detection task because the DA approach assumes that source and target domains share the same feature and output spaces \cite{daume2006domain}.

Although the textual feature itself, along with the existing methods, cannot fully solve this problem, the abundant \textbf{user behavior information} (e.g., browse and purchase information) can offer important potential for cross-domain aspect category detection. Unlike review text, user behavior information can be domain-independent. For instance, a fashionable customer may pay more attention to appearance, style, and similar aspect information, regardless of  domains of the purchased products; while a pragmatist may focus on the quality- and price-related aspects of the target products. This observation motivates us to propose a novel transfer model by leveraging user behavior information, which can be used as intermediate paths for domain integration. 


In this study, we proposed a novel solution, \textbf{Traceable Heterogeneous Graph Representation Learning (THGRL)}, for cross-domain aspect category knowledge transfer and the improvement of review representation learning. By using \textbf{e-commerce user behavior information}, one can connect different domains' textual review information and construct a novel cross-domain heterogeneous graph. Then, supplementary aspect information (knowledge) can random walk from one domain to another. Meanwhile, to efficiently characterize knowledge transfer on the graph, we introduce an innovative latent variable $\mathbb{T}$ (we call it ``Walker Tracer'') to capture global semantic dependencies in the graph. Due to the randomness of the random walk-based generation mechanism, there could be the existence of a lot of noise in graph generated paths (vertex sequences), making the fixed context window based approach problematic. For instance, although an unlabeled review can be walked to a related aspect, the distance (in the walking path) between these two vertexes could be very long. The proposed ``Walker Tracer'' $\mathbb{T}$ aims to summarize, organize, and navigate the graph generated path collections. The basic goal of $\mathbb{T}$ is to find the global semantic coherency pattern across different types of vertexes and relations while eliminating the noisy information. From an algorithm viewpoint, the proposed method can project vertexes (on the heterogeneous graph) and tracers (latent variables on global level) into a low-dimensional joint embedding space. Unlike prior graph embedding approaches \cite{perozzi2014deepwalk,tang2015line,kipf2017semi,jiang2018mathematics}, which focus more on local graph structure representation, the proposed algorithm can capture global random walk patterns across heterogeneous vertexes (customer, review, aspect, and item, etc.). Meanwhile, it is fully automatic without handcrafting feature usage like \cite{dong2017metapath2vec}.

By using THGRL, the sparse (labeled) aspect information can be deliberately diffused to the most related vertexes based on the local/global semantic plus topology information for cross-domain aspect category transfer and detection. 

\textbf{The contribution of this paper is four-fold.}

$\bullet$ First, we propose a novel cross-domain aspect transfer problem for aspect category detection in which the feature spaces, data distributions, and output spaces (of the source and target domains) can be all different. 

$\bullet$ Second, we investigate user behavior information and review textual information to construct a heterogeneous graph, which can enhance aspect category detection performance and address the cold-start problem. 

$\bullet$ Third, an innovative model, Traceable Heterogeneous Graph Representation Learning (THGRL) is proposed to address the cross-domain aspect transfer problem. THGRL can not only auto-embed the graph heterogeneity (i.e., various types of information) but also characterizes the global graphical pattern plus topological information for representation learning. By using THGRL, we project different domains' feature spaces into a common one while allowing data distributions and output spaces stay differently.

$\bullet$ Last but not least, we validate the proposed method by experiments on multiple real-world e-commerce datasets. In order to help other scholars reproduce the experiment outcome, we release the datasets via GitHub\footnote{https://github.com/lzswangjian/THGRL}. To the best of our knowledge, these are the first aspect category detection datasets associated with user behavior information.
\section{Problem Formulation}\label{sec:problem}
\begin{myDef}
\textbf{Aspect Category Detection}
\end{myDef}
For a given product domain $D$ with its own feature space $\mathbb{X}$, an aspect category detection task can be defined as a multi-label classification problem, which is defined by two parts, an output space $ \mathbb{Y}$ and a classification function $F_{c}(\cdot)$. A review instance can be presented as a feature vector $X_{r} = \left \{  x_{1},...x_{N} \right \} \in \mathbb{X}$ where $x_{n}$ denotes $n^{th}$ feature and $N$ is the feature dimension. $ \mathbb{Y}$ is the set of predefined aspect categories (labels); $Y_{r} = \left \{  y_{1},...y_{K} \right \} \in \mathbb{Y}$ represents a binary vector (assigning a value of 0 or 1 for each element); $y_{k}$ denotes $k^{th}$ aspect category; and $K$ is the number of aspect categories. $F_{c}(\cdot)$ is learned from a training set $\Gamma =   \left \{ \left \{ X_{1}, Y_{1} \right \}, ... , \left \{ X_{L}, Y_{L} \right \} \right \}$; each element of $\Gamma$ is a pair of feature vector and label vector;  and $L$ is the training sample number.

\begin{myDef}
\textbf{Cross-Domain Aspect Category Transfer and Detection}
\end{myDef}
$D_{T}$ is the target product domain with its feature space $\mathbb{X}_{T}$. In $D_{T}$, there is a labeled training set $\Gamma_{T} = \left \{ X_{l}, Y_{l} \right \} _{l=1}^{L}$ and an unlabeled data set $\Lambda_{T} = \left \{ X_{ul} \right \} _{ul=1}^{UL}$, $L << UL$ (labeled sample size can be very small). The cross-domain aspect category transfer and detection aim to improve the performance of target classification function $F_{C}^{T}(\cdot)$ by introducing an auxiliary source product domain $D_{S}$.

In this study, we assume:
\begin{itemize}
\item $\mathbb{X}_{T} \neq \mathbb{X}_{S}$. The feature spaces of target and source domains can be different.
\item $\mathbb{Y}_{T} \neq \mathbb{Y}_{S}$. The output spaces (i.e., aspect categories) of target and source domains can be different.
\item $P(Y_{T}|X_{T}) \neq P(Y_{S}|X_{S})$. The data distributions of target and source domains can be different.
\end{itemize}

In this study, we aim to investigate a novel method to project the different $\mathbb{X}_{T}$ and $\mathbb{X}_{S}$ into a common joint graph/text embedding space while reducing the negative effect of different $\mathbb{Y}$ and $P(Y|X)$ by cross-domain knowledge transferring. The proposed method focuses on enhancing heterogeneous graph representation. A more detailed method will be introduced in Section \ref{sec:method}.

\begin{myDef}
\textbf{Heterogeneous Graph}
\end{myDef}
Following the works \cite{sun2011pathsim,dong2017metapath2vec}, heterogeneous graph, namely heterogeneous information network, is defined as a graph $G=(V,E,\tau, \gamma )$ where $V$ denotes the vertex set and $E\subseteq  V\times V$ denotes the edge (relation) set. $\tau$ is the vertex type mapping function, $\tau : V \rightarrow \mathbb{N}$ and $\mathbb{N}$ denotes the set of vertex types. $\gamma$ is relation type mapping function, $\gamma : E \rightarrow \mathbb{U}$ and $\mathbb{U}$ denotes the set of relation types. $\left | \mathbb{N} \right | + \left | \mathbb{U} \right | > 2 $.

\section{Methodology}\label{sec:method}

\subsection{Cross-domain E-commerce Heterogeneous Graph}\label{ssec:graph}

In this study, by leveraging e-commerce behavior information, we enrich the review textual content representation via various types of relations on the heterogeneous graph, which enables cross-domain sentiment aspect transfer and graphical aspect category augmentation. For instance, following a sequence of a customer's purchase and review writing behaviors, the knowledge of a source domain review's aspect categories can be diffused to another unlabeled review in the target domain.

As Table \ref{tab:graph} shows, six types of objects (vertexes) and eight types of relations (edges) are encapsulated in the proposed heterogeneous graph. Note that the vertexes of two different product domains could be shared. For instance, a ``seller'' may sell shoes and clothes simultaneously, and a ``customer'' can also purchase products from both domains. Furthermore, the reviews of two domains may contain the same ``words.'' Meanwhile, they may also share the same aspect categories, such as ``logistics,'' ``seller's services,'' etc. The graph enables cross-domain aspect knowledge transfer.

\begin{table}[htb]
\small
\centering
\caption{Vertexes and edges in the heterogeneous graph}
\label{tab:graph}
\begin{tabular}{| l | l | l |l |}
\hline
\textbf{Vertex}       & \textbf{Description}  &\textbf{Vertex}       & \textbf{Description}                                     \\ \hline
$A$           &  Aspect Category & $R$           &  Review\\ \hline
$W$           &  Word &   $P$           &  Product  \\ \hline
$C$           &  Customer  &  $S$           &  Seller (Online Store)      \\ \hline
\textbf{Edge}       & \multicolumn{3}{l|}{\textbf{Description}}                                        \\ \hline
$P \overset{rec}{\rightarrow} R$           &  \multicolumn{3}{l|}{Product receives a review} \\ \hline
$S\overset{get}{\rightarrow} R$           &  \multicolumn{3}{l|}{Seller gets a review}                            \\ \hline
$C\overset{wri}{\rightarrow} R$           &  \multicolumn{3}{l|}{Customer writes a review}       \\ \hline
$C \overset{pur}{\rightarrow} P$           &  \multicolumn{3}{l|}{Customer purchases a product}                                        \\ \hline
$R \overset{men}{\rightarrow} A$           &  \multicolumn{3}{l|}{Review mentions an aspect category}                             \\ \hline
$W \overset{rel}{\rightarrow} A$          &  \multicolumn{3}{p{6.3cm}|}{Word is related to an aspect category (The word is contained in a review which mentions an aspect category)}   \\ \hline
$R \overset{con}{\rightarrow} W$          &  \multicolumn{3}{l|}{Review contains a word}   \\ \hline
$W \overset{coo}{\rightarrow} W$          &  \multicolumn{3}{l|}{Words co-occur}   \\ \hline
\end{tabular} 
\end{table}


\subsection{Traceable Heterogeneous Graph Representation Learning}\label{ssec:frame}

\begin{algorithm}[htbp]
\caption{Traceable Heterogeneous Graph Representation Learning (THGRL)} 
\label{alo:THGRL}
\begin{algorithmic}[1]
\State \textbf{RepresentationLearning} (Heterogeneous Graph $G=(V,E,\tau, \gamma )$, Dimensions $d$, Walks per Vertex $r$, Walking Length $\zeta$, Context Window Size $cw_{s}$, Walker Tracer Size $|\mathbb{T}|$)
\State Initialize \textit{V\_SEQs} to Empty
\For{\textit{iter = 1} to $r$}
    \ForAll{vertexes $v \in V$}
        \State \textit{V\_SEQ} = \textbf{HierarchicalRandomWalk} ($G,v,\zeta$)
        \State \textbf{Append} \textit{V\_SEQ} to \textit{V\_SEQs}
    \EndFor
\EndFor
\State $\mathbb{T}=$ \textbf{WalkerTracerCapturing} (\textit{V\_SEQs},$|\mathbb{T}|$)
\State $\overrightarrow{f}=$ \textbf{TraceableSkipGram} ($cw_{s},d,\mathbb{T}$,\textit{V\_SEQs})
\State \Return $\overrightarrow{f}$
\\\hrulefill
\State \textbf{HierarchicalRandomWalk} (Heterogeneous Graph $G=(V,E,\tau, \gamma )$, Start vertex $v$, Walking Length $\zeta$)
\State Initialize \textit{V\_SEQ} to $\left \{ v \right \}$
\For{$walk\_step=1$ to $\zeta$}
    \State{Probabilistically draw a relation type $\mathbb{U}_{e}$ based on $v$'s vertex type $\mathbb{N}_{n}$}
    \State{Based on the selected $\mathbb{U}_{e}$, probabilistically draw one vertex  $v_{t}$ from specific relation transition distribution}
    \State{Append $v_{t}$ to \textit{V\_SEQ}}
\EndFor
\State \Return \textit{V\_SEQ}
\end{algorithmic}
\end{algorithm}

To better learn the representation of the constructed e-commerce heterogeneous graph, we propose a Traceable Heterogeneous Graph Representation Learning (THGRL) model. The proposed model consists of three main components: a \textit{hierarchical random walk generator}, a \textit{walker tracer capture mechanism}, and a \textit{vertex-tracer representation co-learning procedure}. The pseudocode for overall algorithm is given in Algorithm \ref{alo:THGRL}.

\textbf{Hierarchical Random Walk}. Unlike the nature of text, a heterogeneous graph ($G=(V,E,\tau, \gamma )$) has more structural and topological characteristics. The vertex's graph neighborhood, $N(v)$, can be defined in various of ways, e.g., direct (one-hop) neighbors of $v$. It is critical to model the vertex neighborhood for graph representation learning. Unlike prior works for homogeneous graph mining, we employ a hierarchical random walk strategy for every vertex $v \in V$ to generate the walking path and vertex embedding. As Algorithm \ref{alo:THGRL} shows, the key step of hierarchical random walk is to sample a relation type $\mathbb{U}_{e}$ from relation type set $\mathbb{U}$. Then, we use the transition distribution of $\mathbb{U}_{e}$ to generate the next move on the graph. A similar hierarchical random walk strategy has proved an effective means to address random walk on the heterogeneous graph \cite{jiang2018cross}. 

\begin{figure}[]\centering
 	\includegraphics[width=1.0\columnwidth]{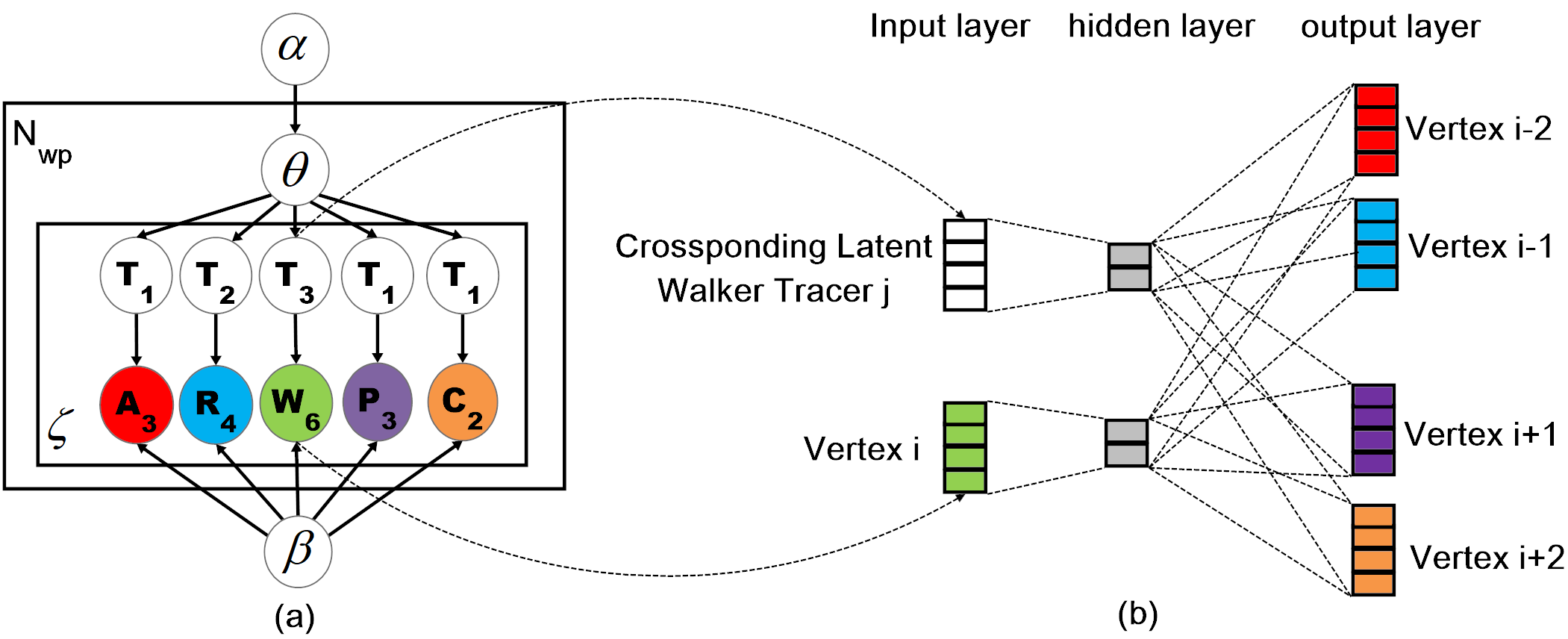}
 	\caption{Walker tracer capturing and vertex-tracer representation co-learning}
 	\label{fig:model}
\end{figure}

\textbf{Walker Tracer Capturing Distribution Learning}. Although the graph heterogeneity has the potential to enhance the random walk performance, the noisy text/behavior information (e.g., noisy vertexes and edges) may pollute the algorithm outcomes. In this work, we propose a novel set of global latent variables, ``Walker Tracer,'' $\mathbb{T} = \left \{ T_{1},...,T_{|\mathbb{T}|} \right \}$ to address this problem. Such variables can capture global random walk dependencies and eliminate the noise.

As Figure \ref{fig:model} (a) shows, in the generated vertex sequences, each vertex can be captured by the corresponding walker tracer(s). A similar method has been utilized to capture latent topics in a given text \cite{blei2003latent, blei2012probabilistic}. We assume the related vertexes tend to appear in the same walking path, and the different appearance patterns of vertexes can be represented as different probability distributions. Such a vertex distribution is a (global) walker tracer. Given a path, the dominant tracer(s) have the higher chance to capture the informative vertex(es) for embedding. Vertex walking paths are then represented as mixtures over these latent tracers. In this study, the following generative process for $N_{wp}$ walking paths of length $\zeta$ with $|\mathbb{T}|$ tracers is defined as: 

\begin{enumerate}
\item For each walking path:
    \begin{enumerate}
        \item Probabilistically draw $\theta_{k} \sim  Dir(\alpha)$
        \item for each vertex in the walking path:
        \begin{enumerate}
            \item Probabilistically draw a specific walker tracer $T_{j} \sim  multi(\theta_{k})$ 
            \item Probabilistically draw a vertex from $Pr(v_{i}|T_{j},\beta)$
        \end{enumerate}
    \end{enumerate}
\end{enumerate}

Here, $\alpha$ is the parameter of the Dirichlet prior on the per-path tracer distributions, and $\beta$ ($Pr(v|T)$) is the vertex capturing distribution for tracer. $\theta$ ($Pr(T|path)$) is the tracer mixture distribution for a walking path.

As mentioned in the introduction, the classical random walk process may bring unexpected noisy information because of the graph heterogeneity. For instance, in a random walk generated path, the aspect-related vertexes could be somehow apart while the context of a vertex can be irrelevant due to the randomness of heterogeneous path generation. This phenomenon could threaten the cross-domain aspect category detection, and existing graph embedding methods \cite{perozzi2014deepwalk,grover2016node2vec,dong2017metapath2vec} can hardly cope with this problem because they can only characterize local random walk information (in a fixed window). In contrast, the proposed global walker tracer is able to estimate the structure of a walk path and capture the long-range informative vertexes.

The probability of vertex-tracer capturing can be calculated as:
\begin{equation}
P(V,\mathbb{T},\theta; \alpha, \beta ) =  \prod_{i=1}^{N_{wp}}Pr(\theta_{i};\alpha) \prod_{j=1}^{\zeta } Pr(\mathbb{T}_{i,j}|\theta_{i})Pr(V_{i,j}|\mathbb{T}_{i,j},\beta)
\end{equation}

The distribution (i.e., $\theta$ and $\beta$) learning is a problem of Bayesian inference. Variational inference \cite{blei2003latent} or Gibbs sampling \cite{porteous2008fast} can be used to address this problem. With the favor of learned distributions, each vertex $v_{i}$ is captured by a latent tracer $T_{j} \in \mathbb{T}$, and the capture probability can be calculated as:

\begin{equation}
Pr(T_{j}|v_{i},path) \propto Pr(v_{i}|T_{j})Pr(T_{j}|path)
\end{equation}

\textbf{Vertex-Tracer Representation Co-Learning}. THGRL obtains the representations of vertexes $V$ (\textit{local information}) and latent walker tracer $\mathbb{T}$ (\textit{global information}) by mapping them into a low-dimensional space $\mathbb{R}^{d}$, $d \ll |V|+|\mathbb{T}|$. The learned representations are able to preserve the topology and semantic information in $G$. Motivated by \cite{liu2015topical}, we propose to learn representations for vertexes and tracers \textbf{\textit{separately}} and \textbf{\textit{simultaneously}}. For each target vertex with its corresponding tracer $\left \langle v_{i},T_{j} \right \rangle$, the objective of THGRL is defined to maximize the following log probability:
\begin{equation}\label{equ:max}
\mathcal{L} = \underset{\overrightarrow{f}}{max}\sum_{v_{i}\in V} \sum_{n \in \mathbb{N}} \sum_{v_{n}^{c} \in N_{n}(v_{i})} [log Pr(v_{n}^{c}|\overrightarrow{f(v_{i})}) + log Pr (v_{n}^{c}|\overrightarrow{f(T_{j})})]
\end{equation}

We use $\overrightarrow{f(\cdot)}$ as the mapping function from multi-typed vertexes and walker tracers to feature representations. Here, $d$ is a parameter specifying the number of dimensions. $N_{n}(v_{i})$ denotes $v_{i}$'s network neighborhood (context) with the $n^{th}$ type of vertexes. As Figure \ref{fig:model} (b) shows, the feature learning method is an upgraded version of the skip-gram architecture, which was originally developed for natural language processing and word embedding \cite{mikolov2013efficient,mikolov2013distributed,bengio2013representation}. Compared with merely using the target vertex $v_{i}$ to predict context vertexes in the original skip-gram model, the proposed approach also employs the corresponding tracer $T_{j}$ of the target vertex. In other words, a tracer is used as a `pseudo vertex' for collective global information representation. So, in the THGRL framework, the vertex's context will encapsulate both local (vertex) and global (tracer) information, which can be critical for cross-domain knowledge transfer and graphical aspect augmentation. 

$Pr(v_{n}^{c}|\overrightarrow{f(v_{i})})$ defines the conditional probability of having a context vertex $v_{n}^{c} \in N_{n}(v_{i})$ given the vertex $v_{i}$'s representation, which is commonly modeled as a softmax function:
\begin{equation}\label{equ:softmax1}
Pr(v_{n}^{c}|\overrightarrow{f(v_{i})}) = \frac{exp(\overrightarrow{f(v_{n}^{c})}\cdot \overrightarrow{f(v_{i})})}{\sum_{u\in V}exp(\overrightarrow{f(u)}\cdot \overrightarrow{f(v_{i})})}
\end{equation}

Similarly, given the representation vertex $v_{i}$'s corresponding tracer $T_{j}$, the conditional probability of having a context vertex $v_{n}^{c} \in N_{n}(v_{i})$ is modeled as:
\begin{equation}\label{equ:softmax2}
Pr(v_{n}^{c}|\overrightarrow{f(T_{j})}) = \frac{exp(\overrightarrow{f(v_{n}^{c})}\cdot \overrightarrow{f(T_{j})})}{\sum_{u\in V}exp(\overrightarrow{f(u)}\cdot \overrightarrow{f(T_{j})})}
\end{equation}

Stochastic gradient ascent is used for optimizing the model parameters of $\overrightarrow{f}$. Negative sampling \cite{mikolov2013distributed} is applied for optimization efficiency. The model parameters size is $(|V|+|\mathbb{T}|)d$ where $d$ is embedding dimension. The computational complexity is $O(IN_{wp}+2cw_{s}N_{wp}(1+\log|V|))$ where $O(IN_{wp})$ indicates the computational complexity of walker tracer capturing distribution learning,  $I$ is the iteration numbers, $N_{wp}$ is walking path numbers; $O(2cw_{s}N_{wp}(1+\log|V|))$ indicates the computational complexity of vertex-tracer representation co-learning, $cw_{s}$ denotes context window size. 

\textbf{Vertex-Tracer Representation Integration}. As aforementioned, walker tracer is defined as a vertex distribution, which indicates the probability of a vertex captured by this tracer. We select the most probable tracer $T_{j}$ of vertex $v_{i}$ to generate the integrated vertex-tracer representation:
\begin{equation}\label{equ:mvtr}
\overrightarrow{f(v_{i}^{T})_{I}}=\mathop{\arg\max}_{T_{j}} Pr(T_{j}|v_{i})\overrightarrow{f(v_{i}^{T_{j}})_{c}}
\end{equation}
where $\overrightarrow{f(v_{i}^{T_{j}})_{c}}$ is the combined representation of vertex $v_{i}$ under tracer $T_{j}$, obtained by concatenating the embedding of $v_{i}$ and $T_{j}$, i.e., $\overrightarrow{f(v_{i}^{T_{j}})_{c}}=\overrightarrow{f(v)} \bigoplus \overrightarrow{f(T_{j})}$, where $\bigoplus$ is the concatenation operation. 



Finally, for the aspect category detection task, given a review instance $R = \left \{ W_{1},...,W_{n} \right \}$, the vertex-tracer representations of all review words are further down-sampled with a global average-pooling operation. The learned review representation can be used as input for classification function $F_{c}(\cdot)$ for training and testing. Meanwhile, as stated above, the proposed method focuses on cross-domain aspect category transfer via enhanced heterogeneous graph representation. Without loss of generality, $F_{C}(\cdot)$, can be a multi-label classification model with an arbitrary structure, i.e., kernel-based model like SVM \cite{steinwart2008support} (for limited training samples) or neural network based model like BiLSTM-Attention~\cite{yang2016hierarchical} (for sufficient training samples). 

With THGRL, objects from different domains can be mapped into a same feature space (addressing the ``various feature spaces'' problem), and the global semantic/aspect dependencies can be characterized for aspect category knowledge transfer (reducing the negative effect of ``different data distributions'' and ``diverse output spaces'' problems).
\section{Experiment}

\subsection{Dataset and Experiment Setting}\label{ssec:data}
\textbf{Dataset}\footnote{To the best of our knowledge, behavior information, e.g., customer purchases and seller sells information, is not publicly available. Meanwhile, all the public datasets do not associate with these kinds of information. In order to address this problem, we collected new datasets (https://github.com/lzswangjian/THGRL) from Taobao.}. In Table \ref{tab:data}, we summarize statistics of three e-commerce datasets (product domains) from Taobao (an online consumer-to-consumer platform in Alibaba). For a specific domain, we collected the customers' purchase behaviors. For each purchase behavior record, the product ID, the seller ID (online store), and the customer ID were collected. Meanwhile, the reviews of the purchased product and the words contained in the target reviews were also collected. The related aspect categories of reviews were manually labeled by a third-party. For this study, we validated the proposed algorithm in four cross-domain aspect category detection tasks: (\textit{clothing$\rightarrow$shoes}, \textit{shoes$\rightarrow$clothing}, \textit{clothing$\rightarrow$bags}, and \textit{bags$\rightarrow$clothing}).

\begin{table}[!htb]
\small
\centering
\caption{Statistics of the e-commerce datasets from Taobao}
\label{tab:data}
\begin{threeparttable}
\begin{tabular}{|l | p{0.9cm} | p{0.9cm} | p{0.9cm} | p{0.9cm} |}
\hline
\multicolumn{5}{|l|}{\textbf{Datasets}} \\\hline
\textbf{Domain}      & \textbf{Shoes} &  \textbf{Bags}& \multicolumn{2}{l|}{\textbf{Clothing} }\\ \hline
\textbf{\#Vertex}  & 18,471 & 25,066 & \multicolumn{2}{l|}{15,330}  \\ \hline
\textbf{\#Vertex Type} & 6  &    6        & \multicolumn{2}{l|}{6} \\ \hline
\multicolumn{5}{|l|}{\textbf{Cross-Domain Aspect Category Detection Tasks (Graph)}} \\\hline
&\multicolumn{2}{l|}{\textbf{Clothing$\rightarrow$Shoes}}&\multicolumn{2}{l|}{\textbf{Clothing$\rightarrow$Bags}} \\
\textbf{Task (Graph)}&\multicolumn{2}{l|}{\textbf{Shoes$\rightarrow$Clothing}}&\multicolumn{2}{l|}{\textbf{Bags$\rightarrow$Clothing}} \\ \hline
\textbf{\#Shared Aspects} & \multicolumn{2}{l|}{8} & \multicolumn{2}{l|}{8}\\ \hline
\textbf{\#Specific Aspects} &10 &7 &10 &9 \\ \hline
\textbf{\#Shared Customers} & \multicolumn{2}{l|}{2,044} & \multicolumn{2}{l|}{186}\\ \hline
\textbf{\#Specific Customers} &5,715 &7,874 & 7,573& 10,420 \\ \hline
\textbf{\#Shared Sellers} & \multicolumn{2}{l|}{23} & \multicolumn{2}{l|}{12}\\ \hline
\textbf{\#Specific Sellers} & 3 & 14 & 14 & 0 \\ \hline
\textbf{\#Shared Words} & \multicolumn{2}{l|}{3,541} & \multicolumn{2}{l|}{3,903}\\ \hline
\textbf{\#Specific Words} & 3,807 & 4,814 & 3,445 & 10,402 \\ \hline
\textbf{\#Specific Reviews$^{2}$} & 	
10,261 & 10,590 & 10,261 & 11,031 \\ \hline
\textbf{\#Specific Products$^{2}$} & 179 & 146 & 179 & 126 \\ \hline
\textbf{\#Edge}      & \multicolumn{2}{l|}{694,036}   &  \multicolumn{2}{l|}{1,134,742}  \\ \hline
\textbf{\#Edge Type} & \multicolumn{2}{l|}{8} & \multicolumn{2}{l|}{8}\\ \hline
\end{tabular}
\begin{tablenotes}
\item * There are no shared reviews and products.
\end{tablenotes}
\end{threeparttable}
\end{table}

\textbf{Baselines and Comparison Groups}. We chose two groups of baseline algorithms, from text and graph viewpoints, to comprehensively evaluate the performance of the proposed method. 

\textbf{\textit{Textual Content Based Baseline Group}}\footnote{In the textual content based group, the SVM model was using \textit{TFIDF} feature vector, and neural network based models were using 300 dimensional dense feature vectors pre-trained by FastText \cite{grave2017bag} in an enormous corpus provided by Taobao.}: this group of baselines only utilized textual information for aspect category detection tasks.

1. Support Vector Machine~\cite{steinwart2008support}: We followed \cite{ganu2009beyond} to train an ``one vs. all'' classifier on target domain reviews. The similar approach had achieved the top ranking in the aspect category detection subtask of SemEval-2014 \cite{kiritchenko2014nrc}. This baseline was denoted as \textbf{SVM}.

2. Convolutional Neural Network~\cite{kim2014convolutional}: We trained the convolutional neural networks (CNN) on top of the pre-trained word vectors for aspect category detection task. \cite{ruder2016insight} utilized this approach for aspect-level sentiment analysis. This baseline was denoted as \textbf{CNN}.

3. Bidirectional Recurrent Neural Network (BiLSTM) with Attention Mechanism~\cite{liu2017attention}: We used a bidirectional LSTM to represent the word sequence in a review then calculated the weighted values over each word in reviews by an attention model. This baseline was denoted as \textbf{BiLSTM-Attention}.

4. Transformer~\cite{vaswani2017attention}: We used an encoder structure of Transformer model for learning the review textual representation. As a state-of-the-art model to encode deep semantic information using self-attention mechanism, a similar process has been utilized in several research works for different downstream tasks \cite{young2018recent}. This baseline was denoted as \textbf{Transformer}.

5. Semi-Supervised Learning: First, an SVM model on limited ``real'' training data was initially trained. Second, 1,000 unlabeled reviews were selected for pseudo labeling by the previous trained model. Third, the pseudo-labeled data was added into training set for model re-training. We repeated this process until the best performing model was finally found. This baseline only used target domain reviews for training, denoted as \textbf{Semi-Supervised}.

6. Domain adaptation~\cite{zellinger2017central}: This recent model is the state-of-the-art domain adaptation method for sentiment classification which learns the domain-invariant representations with neural networks. This baseline requires the training review data from both source and target domains, denoted as \textbf{Domain Adaptation}. We tuned the weighting parameter $\lambda $, and picked up a best performed parameter setting for experiment.

\textbf{\textit{Graph Embedding Based Baseline Group}}\footnote{All graph based baseline algorithms employed SVM as a classification model for aspect category detection tasks. The input representation was obtained by concatenating the \textit{TFIDF} feature vector and graph embedding feature vector.}: This group of baselines generated the graphical embeddings based on the constructed heterogeneous graphs which integrated multiple types of information (including textual information and user behavior information).

7. DeepWalk~\cite{perozzi2014deepwalk}: We used a DeepWalk algorithm to learn the graph embeddings, denoted as \textbf{DeepWalk}.

8. LINE~\cite{tang2015line}: This model was aimed at preserving first-order and second-order proximity in concatenated embeddings, denoted as \textbf{LINE}.

9. Node2vec~\cite{grover2016node2vec}: We used node2vec algorithm to learn graph embeddings via second order random walks in the graph, denoted as \textbf{Node2vec}. We tuned return parameter $p$ and in-out parameter $q$ with a grid search over $p, q \in \left \{ 0.25, 0.50, 1, 2, 4 \right \}$ and picked up a best performing parameter setting for the experiment, as suggested by~\cite{grover2016node2vec}.

10. Metapath2vec++~\cite{dong2017metapath2vec}: This model was originally designed for heterogeneous graphs. It learns heterogeneous graph embeddings via metapath based random walk and heterogeneous negative sampling in the graph. Metapath2vec++ requires a human-defined metapath scheme to guide random walks. We tried two different metapaths for this experiment: (1) $R \overset{con}{\rightarrow} W \overset{rel}{\rightarrow} A$ (this metapath was associated with textual information solely, denoted as \textbf{Metapath2Vec++(T)}), (2) $C \overset{pur}{\rightarrow} P \overset{rec}{\rightarrow} R \overset{get}{\leftarrow} S \overset{get}{\rightarrow} R \overset{con}{\rightarrow} W \overset{rel}{\rightarrow} A \overset{rel}{\leftarrow} R$ (this metapath was not only related to the textual information, but also involving behavior information, denoted as \textbf{Metapath2Vec++(B)}).
 
11. Graph Convolutional Networks~\cite{kipf2017semi}: We trained convolutional neural networks on an adjacency matrix of graph and a topological feature matrix of vertexes in a vertex classification task (semi-supervised task), as suggested by~\cite{kipf2017semi}. This more recent baseline was denoted as \textbf{GCN}.

Please note that, because DeepWalk, LINE, Node2vec, and GCN were originally designed for homogeneous graph, for a fair comparison, we have to construct a homogeneous graph based on the heterogeneous one. We first integrated all relations between two vertexes into one edge, then estimated the edge weight (transition probability) by summing of all integrated relations.

\textbf{\textit{Comparison Groups}}: We compared the performances of several variants of the proposed method in order to highlight our technical contributions. 

\textbf{THGRL}\bm{$_{Def}$}: The default setting for the proposed THGRL model, which used hierarchical random walk generator for walking path generation, integrated vertex-tracer representation for graph embedding, and SVM model for aspect category detection task. 

\textbf{THGRL}\bm{$_{OV}$}: We removed the walker tracer representation information from vertex-tracer representation (only vertex representation left). 

\textbf{THGRL}\bm{$_{Ran}$}: We replaced the hierarchical random walk generator by an ordinary random walk generator. 

\textbf{THGRL}\bm{$_{NN}$}: We replaced the SVM classification model by a BiLSTM-Attention based classification model.

\begin{table*}[]
\footnotesize
\centering
\caption{Results of cross-domain aspect category detection tasks}
\label{tab:result}
\begin{threeparttable}
\begin{tabular}{|l|l|l|l|l|l|l|l|l|l|}
\hline
\multicolumn{2}{|l|}{\multirow{2}{*}{\textbf{Method}}} & \multicolumn{2}{l|}{\textbf{Clothing$\rightarrow$Shoes}} & 
\multicolumn{2}{l|}{\textbf{Shoes$\rightarrow$Clothing}} &
\multicolumn{2}{l|}{\textbf{Clothing$\rightarrow$Bags}} &
\multicolumn{2}{l|}{\textbf{Bags$\rightarrow$Clothing}}\\
\multicolumn{2}{|l|}{}& \multicolumn{2}{l|}{\textbf{(Train:70; Test:8397)}} & 
\multicolumn{2}{l|}{\textbf{(Train:100; Test:6972)}} &
\multicolumn{2}{l|}{\textbf{(Train:90; Test:5523)}}  &
\multicolumn{2}{l|}{\textbf{(Train:100; Test:6972)}} \\ \hline
\multirow{6}{*}{\begin{tabular}[c]{@{}l@{}}\textbf{Text}\end{tabular}} &\textbf{Textual Content Based Baseline Group}& \textbf{Micro-F1} & \textbf{Macro-F1} &  \textbf{Micro-F1} & \textbf{Macro-F1} & \textbf{Micro-F1} & \textbf{Macro-F1} & \textbf{Micro-F1} & \textbf{Macro-F1}\\\cline{2-10}
&SVM~\cite{ganu2009beyond}& 0.4255 & 0.2938 & 0.4736& 0.3636& 0.2045  & 0.1982 & 0.5210 & 0.3700\\
&CNN~\cite{kim2014convolutional}& 0.3056  & 0.1457 & 0.2588 & 0.1529 &  0.2463 & 0.1529 &  0.2830 & 0.1632\\
&BiLSTM-Attention~\cite{liu2017attention}& 0.4239 & 0.2520  & 0.4357& 0.2601 & 0.3349 & 0.2652 & 0.4094 & 0.2464\\ 
&Transformer~\cite{vaswani2017attention}& 0.3997 & 0.2106 & 0.3447 & 0.2252 & 0.3183 & 0.2699 & 0.3456 & 0.2113\\
&Semi-Supervised & 0.4321 & 0.3604 & 0.4842 & 0.3706
& 0.3439  & 0.2688 & 0.4213 & 0.3445\\
&Domain Adaptation~\cite{zellinger2017central} & 0.4700  & 0.3298 &0.5525 & 0.4429 & 0.3412  & 0.2931  & 0.5043 & 0.4252 \\\hline
\multirow{13}{*}{\begin{tabular}[c]{@{}l@{}}\textbf{Graph$^{1}$}\end{tabular}} &\textbf{Graph Embedding Based Baseline Group}& \textbf{Micro-F1} & \textbf{Macro-F1} &  \textbf{Micro-F1} & \textbf{Macro-F1} & \textbf{Micro-F1} &  \textbf{Macro-F1} & \textbf{Micro-F1} &  \textbf{Macro-F1} \\\cline{2-10}
& DeepWalk~\cite{perozzi2014deepwalk}& 0.4992 & 0.3858 & 0.6295& 0.5592 & 0.3730 & 0.3599 & 0.6332 &  0.5291
 \\
& LINE~\cite{tang2015line} & 0.5057 & 0.3950  & 0.6489 & 0.5646& 0.3848  & 0.3603 & 0.6293 & 0.5223 \\
& Node2vec~\cite{grover2016node2vec} & 0.4858 & 0.3658 & 0.6308& 0.5701 & 0.3787 & 0.3678 & 0.6326 & 0.5498 \\
& Metapath2Vec++(T)~\cite{dong2017metapath2vec} & 0.4574 & 0.3854 &0.6702& 0.5477&  0.3573  & 0.3401 & 0.5360 & 0.4387 \\
& Metapath2Vec++(B)~\cite{dong2017metapath2vec} & 0.4514 & 0.3302 & 0.5444 & 0.4199 & 0.2360  & 0.2343  & 0.5678 & 0.4133\\
& GCN~\cite{kipf2017semi} & 0.4424  & 0.3242 & 0.5027& 0.3840 & 0.2116  & 0.2176 & 0.5411 & 0.3841 \\\cline{2-10}
& \textbf{Comparison Group}  & \textbf{Micro-F1}  & \textbf{Macro-F1}  & \textbf{Micro-F1}  & \textbf{Macro-F1}  & \textbf{Micro-F1}  & \textbf{Macro-F1} & \textbf{Micro-F1}  & \textbf{Macro-F1} \\ \cline{2-10}
& THGRL$_{OV}$ & 0.5167  & 0.4138 & 0.6721 & 0.5913 & 0.4039  & 0.3858 & 0.6479 & 0.5827 \\
& THGRL$_{Ran}$  & 0.5307  & 0.4160  & 0.6550 & 0.5644 & 0.4109  & 0.3844 & 0.6228 & 0.5123\\
& THGRL$_{NN}$  & 0.4668  & 0.2896 & 0.4864 & 0.3366
 & 0.4032 &  0.3503 & 0.5461 & 0.3526 \\
& \textbf{THGRL}\bm{$_{Def}$} \textbf{(Proposed Method)}   & \textbf{0.5376*}  & \textbf{0.4381*} & \textbf{0.6982*} & \textbf{0.6122*} &  \textbf{0.4244*} & \textbf{0.3951*} & \textbf{0.6579*} & \textbf{0.5920*} \\\hline
\end{tabular}
\begin{tablenotes}
\item[1] Unless otherwise noted, all graph-based algorithms (including baseline group and comparison group) were using SVM as classification model, and the input representation was obtained by concatenating the \textit{TFIDF} feature vector and graph embedding feature vector.
\end{tablenotes}
\end{threeparttable}
\end{table*}

\textbf{Training and Testing Set}. An additional source domain may bring many more training samples for shared aspect categories. For a fair comparison and better addressing the research questions, we only detected the target domain-specific aspect categories. For all labeled review instances in the target domain, limited training samples were randomly selected (cold start). Each target domain-specific aspect category would have only 10 labeled review samples for training, while the rest of the reviews were used as the testing set (all $W \overset{rel}{\rightarrow} A$ and $R \overset{men}{\rightarrow} A$ relations from the testing set were removed for fair comparison). This experimental setting can be very challenging. For instance, in the \textit{clothing $\rightarrow$ shoes} task, the target domain featured seven specific aspects, and there were only 70 training instances (the testing instance number was 8,397). The classification models (including baselines and proposed model) were trained on labeled data from the target domain. The only exception was the baseline model ``Domain Adaptation'' \cite{zellinger2017central}, which required training data from both domains. For evaluation, the different models were evaluated by two commonly used measures, Micro-F1 and Macro-F1.

\textbf{Experimental Set-up}. For the proposed THGRL model, we utilized the following setting: (1) the number of walks per vertex $r$: 10; (2) the walk length $\zeta$: 80; (3) the embedding dimension $d$: 128; (4) the context window size $cw_{s}$: 10. For experiment fairness, all the random walk based embedding baselines shared the same parameters. Please note that we didn't tune those parameters. Most graph-embedding methods reported the above parameter settings in their original paper \cite{perozzi2014deepwalk,grover2016node2vec}. The walker tracer number was 100.

\subsection{Experiment Result and Analysis}

\textbf{Overall Results}. The cross-domain aspect category detection performance results of different models are reported in Table \ref{tab:result}. Based on the experiment results, we have the following observations: 

$\bullet$ \textbf{THGRL vs. Baselines.} The proposed method outperformed the other baseline models for all evaluation metrics in all tasks. For instance, in terms of Micro-F1, THGRL$_{Def}$ outperformed the best-performing baseline by 10.3\% for ``\textit{clothing $\rightarrow$ bags}'' task. With respect to Macro-F1, THGRL$_{Def}$ outperformed the best-performing baseline by 11.0\% for ``\textit{clothing $\rightarrow$ shoes}'' task.

$\bullet$ \textbf{Graph vs. Text.} The baselines (e.g., SVM~\cite{ganu2009beyond}, CNN~\cite{kim2014convolutional}, Transformer~\cite{vaswani2017attention}, and BiLSTM-Attention~\cite{liu2017attention}) solely relied on textual information and didn't perform well. Correspondingly, most graph-embedding approaches with different mechanisms can improve task performance. This observation proves the hypothesis that user behavior information can provide important potentials for cross-domain aspect category transfer and detection.

$\bullet$ \textbf{Deep vs. Simple.} Unfortunately, the deep neural network approach in this experiment cannot outperform the simple classification models, mainly because of the training data sparsity. As section \ref{sec:intro} mentioned, the goal of this work is to address the training data sparseness problem in the target domain (e.g., 70 training instances and 8,397 testing instances for $Clothing\rightarrow Shoes$ task). Compared with the simple models, the deep learning family needs more training data for optimization.

$\bullet$ \textbf{Different Mining Methods on Textual Information.} By introducing more pseudo-labeling data, the semi-supervised approach could somehow improve the detection performance. Meanwhile, domain adaptation model~\cite{zellinger2017central} was proposed to learn the invariant knowledge from both domains, which could address the data distribution difference. However, this model cannot efficiently cope with the feature and aspect (label) difference of the source and target domains. The performances of these two baselines were still unsatisfactory. This phenomenon shows, when the labeled samples are quite limited, the marginal effect that can be obtained by mining the textual information is also limited.

$\bullet$ \textbf{Cross Domain vs. Single Domain.} The experimental results of the domain adaptation model and all graph embedding-based models indicate that an additional related source domain can be useful for the target domain's task. In more detail, most graph embedding-based models (including baselines and comparison group) can outperform the domain adaptation model (only using textual information). This observation further proves the usefulness of user behavior information.

$\bullet$ \textbf{Metapath2Vec++.} Although designed for heterogeneous graph embedding, two Metapath2Vec++~\cite{dong2017metapath2vec} baselines didn't show a significantly different performance from other homogeneous graph embedding models. A possible explanation is that, in this task, no single metapath can cover the aspect detection requirement. In addition, metapath based random walk can be too strict to explore potential useful neighbourhoods for graph representation learning.

$\bullet$ \textbf{GCN.} GCN~\cite{kipf2017semi} didn't perform well in the experiment. The reason may be multifaceted. First, GCN is designed for a homogeneous graph, but in this study, we utilized heterogeneous graphs. Second, the semi-supervised task of original GCN model was vertex classification, which was inconsistent with final task.

$\bullet$ \textbf{Components of THGRL.} To evaluate the components of the proposed method: (1) when we removed the tracer representation information from vertex-tracer representation, the aspect detection performance decreased in all tasks. This result shows that the proposed ``walker tracer'' does  capture the useful global semantic information, which plays an important role in eliminating noise in graph random walks and enhances the task performance. (2) If we replaced the hierarchical random walk generator with an ordinary random walk generator, the performance also declined in all tasks. It is clear that hierarchical random walk can contribute to the heterogeneous graph based random walk and graph representation accuracy significantly. (3) because of the sparseness of the training data, a relatively simple classification model can outperform the sophisticated neural network based models.

\begin{figure}[]\centering
 	\includegraphics[width=1.0\columnwidth]{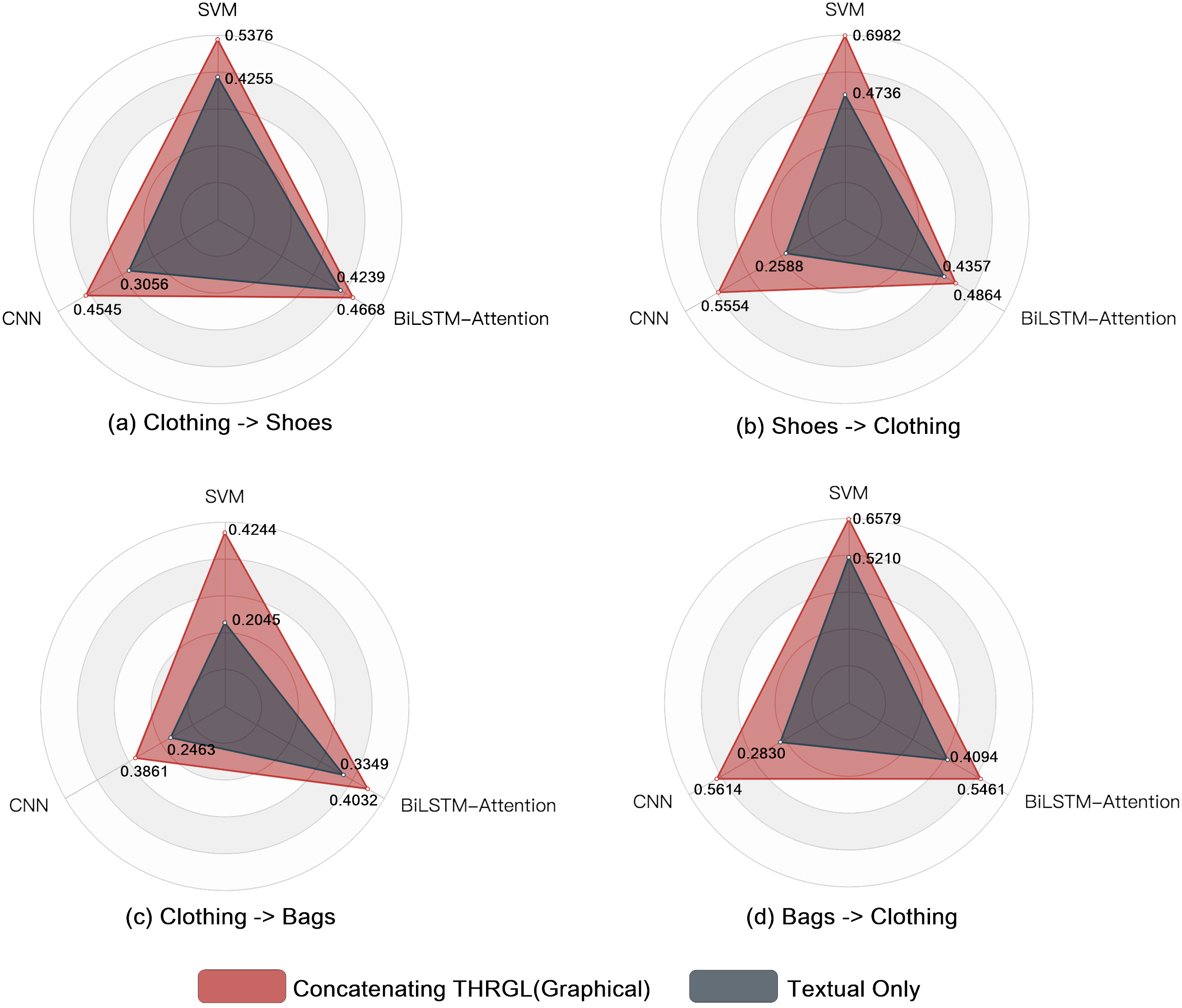}
 	\caption{Aspect category detection performance (Micro-F1) using different feature vectors, for multiple classification models, in (a) ``clothing $\rightarrow$ shoes'' task (train:70; test:8397); (b) ``shoes $\rightarrow$ clothing'' task (train:100; test:6972); (c) ``clothing $\rightarrow$ bags'' task (train:90; test:5523); (d) ``bags $\rightarrow$ clothing'' task (train:100; test:6972).}
 	\label{fig:feature}
\end{figure}

\textbf{Comparison of Different Feature Vectors with Various Classification Models}. To gain a deeper understanding regarding the representation capacity of the proposed THGRL method, we compared the aspect category detection task performance using different classification models with different features. As Figure \ref{fig:feature} shows: (1) compared to solely using the textual information (\textit{TFIDF} feature vectors for SVM and pretrained dense word feature vectors for neural network based models), by utilizing ``THGRL'' embedding, all task performances show significant improvements. For instance, the CNN classification model can achieve an improvement up to 115\% (from 0.2588 to 0.5554) in the ``shoes $\rightarrow$ clothing'' task; while the SVM classification model can achieve an improvement up to 107\% (from 0.2046 to 0.4244) in the ``clothing $\rightarrow$ bags'' task. (2) the SVM model with THGRL features achieved the best performance in all tasks. This observation once again confirms that a relatively simple model can outperform the complicated models in the case of limited training samples. (3) Overall, the CNN model can achieve the greatest improvement (an average increase of 79.6\%). The SVM model comes second (an average increase of 51.9\%). The improvement of BiLSTM-Attention is relatively small (an average increase of 18.9\%). 

The above comparison results also demonstrate that user behavior information is vital in the aspect category detection task, and the proposed method could learn a better representation to effectively accomplish the cross-domain transfer and information augmentation tasks.

\begin{figure}[]\centering
 	\includegraphics[width=1.0\columnwidth]{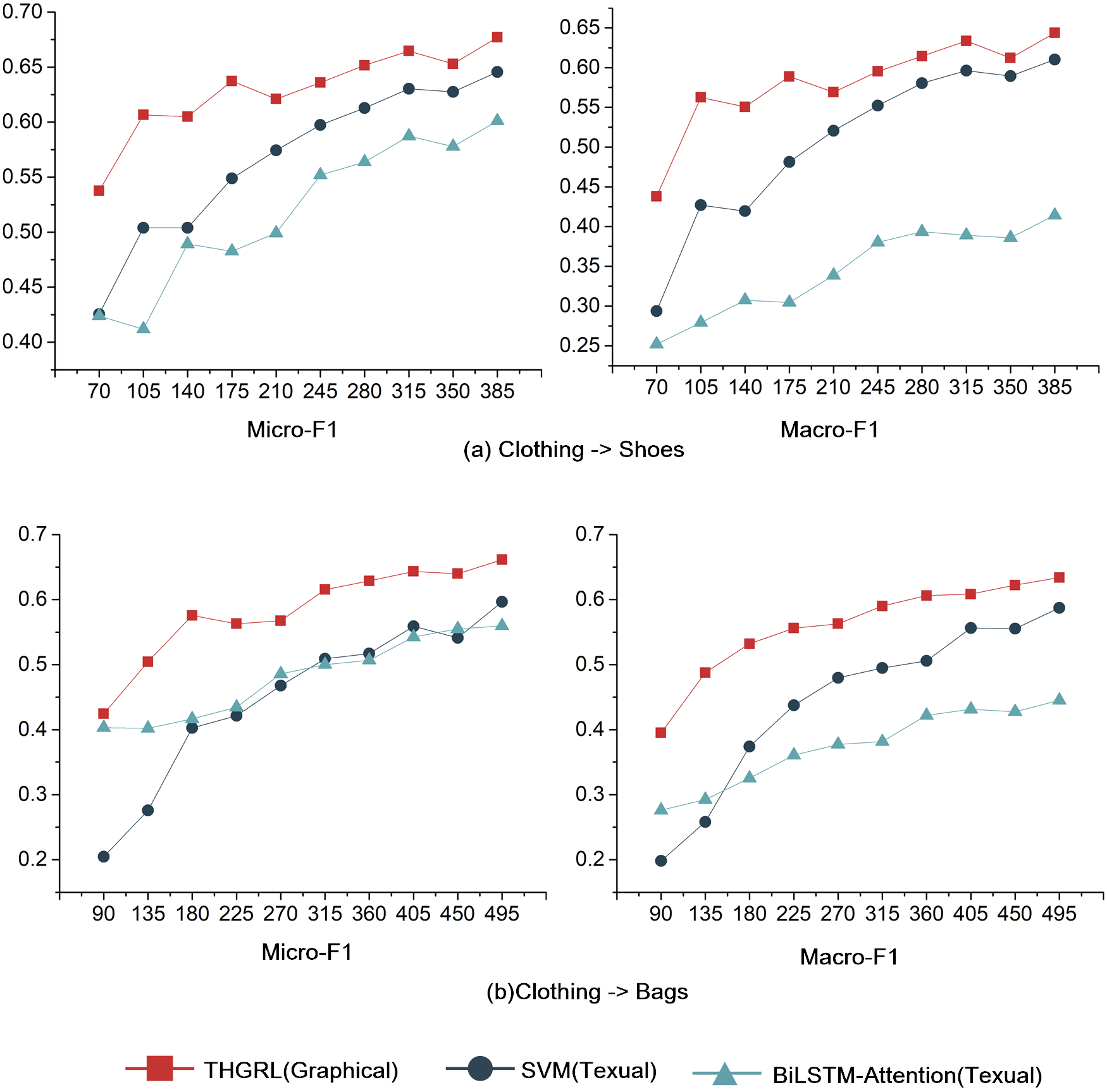}
 	\caption{Aspect category detection performance based on increasing training samples (tracer:100, dimension:128) of  (a)``clothing $\rightarrow$ shoes'' task; (b) ``clothing $\rightarrow$ bags'' task}
 	\label{fig:increase}
\end{figure}

\begin{figure}[]\centering
 	\includegraphics[width=0.9\columnwidth]{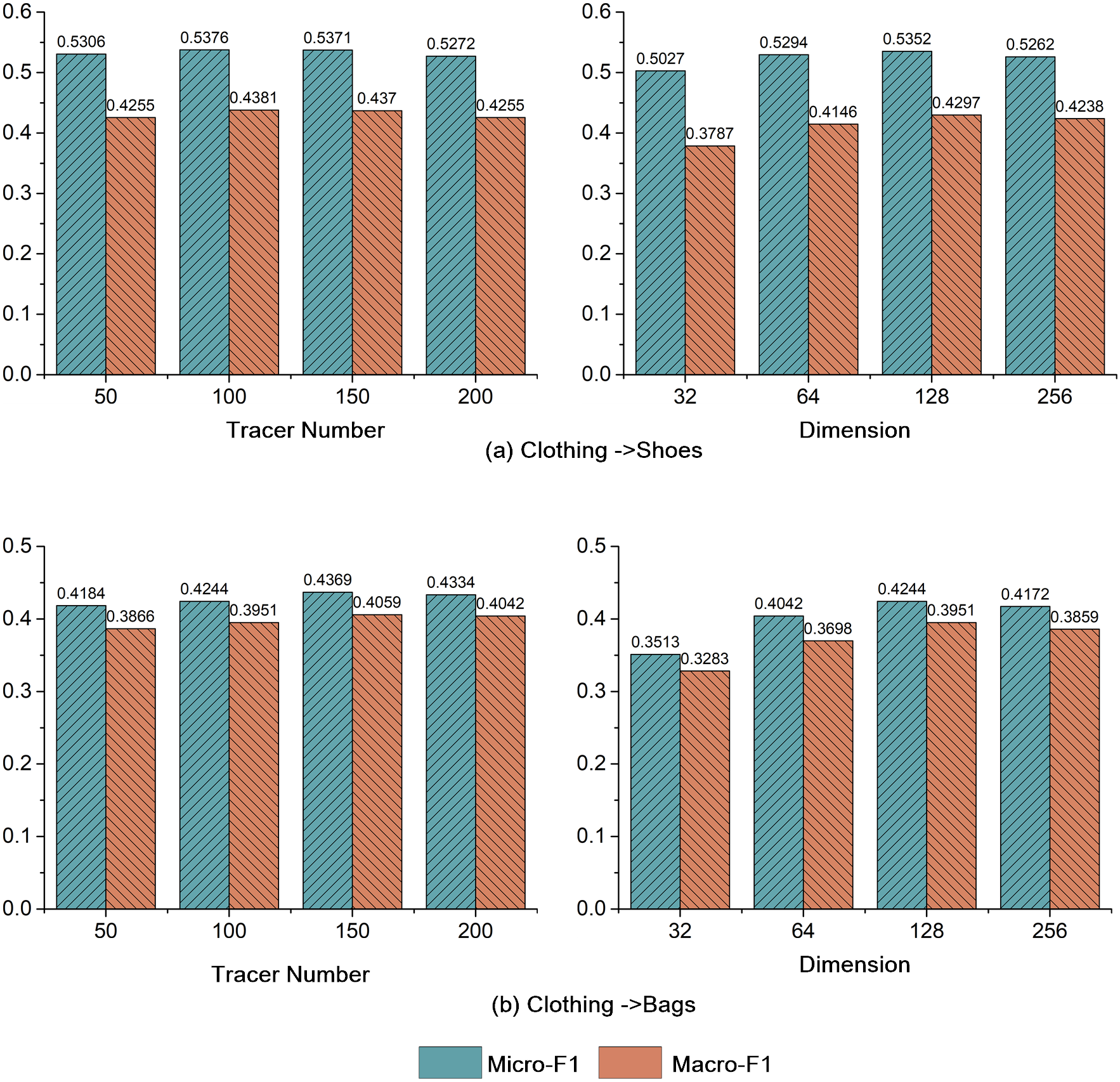}
 	\caption{Aspect category detection performance under different tracer numbers and representation dimensions in (a) ``clothing $\rightarrow$ shoes'' task; (b) ``clothing $\rightarrow$ bags'' task.}
 	\label{fig:parameter}
\end{figure}

\textbf{Trends under Increasing Training Samples}. To further validate the performance of the proposed method, we compared the aspect category detection performance by continually adding training samples in two cross-domain tasks. As Figure \ref{fig:increase} shows: (1) By adding more training samples, the performance of all models have improved. The proposed method is consistently better than SVM and BiLSTM-Attention which only employing textual features. (2) When the training samples increase, the gap between the THGRL and SVM narrows. (3) The performance of BiLSTM-Attention is not very stable. For instance, in the ``clothing $\rightarrow$ shoes'' task, it performs worst in most cases, while in the ``clothing $\rightarrow$ bags'' task, the growth rate of this model is relatively small.

\textbf{Parameter Sensitivity Analysis}
We also conducted a sensitivity analysis of THGRL by tailoring the tracer number and representation dimensions. Figure \ref{fig:parameter} depicts their impacts on the aspect category detection performance. Based on the comparison, we find that, in the ``clothing $\rightarrow$ shoes'' task, the proposed method was not very sensitive to these two hyper-parameters, while 100 tracers and 128 dimensions were the best-performing parameter setting. Meanwhile, in the ``clothing $\rightarrow$ bags'' task, 150 tracers and 128 dimensions were the best-performing parameter setting, and the change of the hyper-parameters had a relatively large impact for task performance. This may be caused by different characteristics of different cross-domain tasks. As shown in Table \ref{tab:data}, in the ``clothing $\rightarrow$ shoes'' task, there are more shared customers and sellers (bridges on the graph). Hence, the graphical representation learning could be easier. Furthermore, the constructed heterogeneous graph of ``clothing $\rightarrow$ bags'' has more vertexes and edges, which may require more tracers to achieve the optimal performance.

\begin{figure}[]\centering
 	\includegraphics[width=1.0\columnwidth]{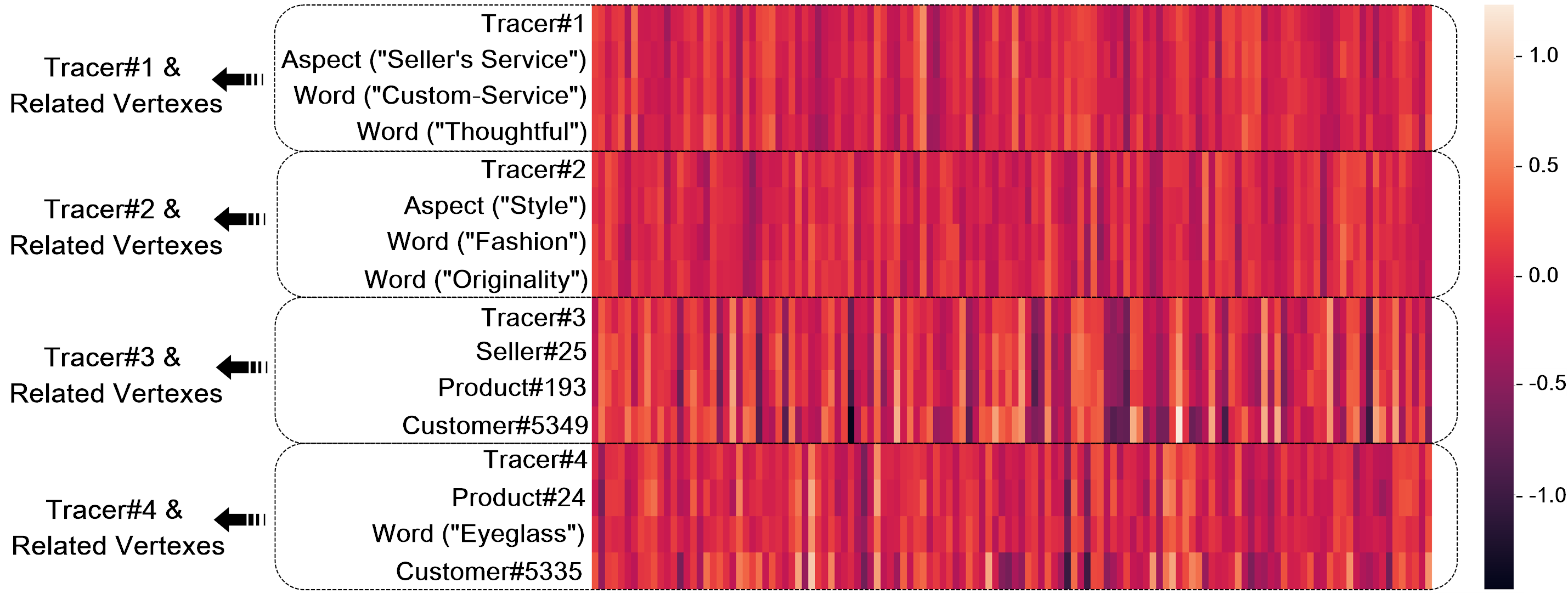}
 	\caption{Heat map of learned traceable heterogeneous graph representation in ``clothing $\rightarrow$ shoes'' task}
 	\label{fig:case}
\end{figure}

\textbf{Embedding Visualization}. We used a heat map to visualize the traceable heterogeneous graph representation of four tracers and the associated vertexes in the experiment. In this heat map, each row is a representation, and colors depict data values. As Figure \ref{fig:case} shows, for each walker tracer, a group of closely related heterogeneous vertexes (e.g., aspects, words, products, customers) are successfully captured. The vertexes in the same group tend to deliver similar semantic knowledge. For instance, the vertexes with high capturing probability of Tracer\#1, are all closely related to service, e.g., aspect vertex of ``seller's service,'' word vertexes of ``custom-service'' and ``thoughtful.'' Meanwhile, different types of vertexes can be captured simultaneously. For instance, the types of vertexes with high capturing probability of Tracer\#4 are very diverse, including product, word and customer, etc.

From the similarity viewpoint, it is clear that the representations (color pattern) between various groups are significantly different; while in the same group, they are very similar.  

These observations indicate that: (1) The e-commercial behavior information (among customer, seller and products) follows certain patterns, and these behavior patterns have the potential to mirror the aspect information for review mining; (2) The proposed traceable heterogeneous graph representation learning approach could successfully capture this kind of global information, which could be used for aspect augmentation and aspect knowledge transfer.
\section{Related Work}\label{review}
\textbf{Aspect-level review analysis}. Aspect-level review analysis is a fine-grained opinion mining task. Identifying the aspect category helps to get target-dependent sentiment and contributes to aspect-specific opinion summarization \cite{zhou2015representation}. Prior works mainly focused on aspect extraction \cite{hu2004mining,titov2008modeling}. Recently, International Workshop on Semantic Evaluation (SemEval) developed a series of tasks with the aim of facilitating and encouraging research in aspect based sentiment analysis and its related fields \cite{pontiki2014semeval,pontiki2016semeval}. Aspect category detection aims to identify the related aspects expressed in a given review, which is a fundamental task for aspect-level sentiment analysis \cite{tang2016aspect}. Ganu et al.~\cite{ganu2009beyond} used SVM to train one vs. all classifiers on restaurant review datasets for aspect category detection. TFIDF vector of stem words was used as features in their study. A similar algorithm was applied in \cite{kiritchenko2014nrc} with a Yelp word-aspect association lexicon to boost the performance. McAuley et al.~\cite{mcauley2012learning} proposed a discriminative model to predict product aspect. With the rise of deep learning, recent works began to adopt neural network structure in their researches. \cite{ruder2016insight} used continuous word representations and a convolutional neural network (CNN) for review representation learning, while \cite{nguyen2015phrasernn} used Recursive Neural Network(RNN). Bi-LSTM with attention mechanism \cite{yang2016hierarchical} was applied in \cite{liu2017attention} for aspect-level sentiment analysis. Although the existing methods have achieved promising performance, they all rely on textual information for aspect-level review analysis. Unlike existing works, in this study, we investigated a novel and important potential of aspect category detection by leveraging user behavior information.

\textbf{Transfer learning}. It's often expensive and time consuming to obtain enough labeled reviews for aspect category detection tasks. Domain adaptation, a.k.a, homogeneous transfer learning \cite{daume2006domain,weiss2016survey}, was proposed for training and testing models on different domain distributions with same feature and output spaces. For instance, Glorot et al.~\cite{glorot2011domain} proposed a deep learning model for large-scale cross-domain sentiment polarity classification. Zellinger et al.~\cite{zellinger2017central} tried to learn the domain-invariant representations in the context of domain adaptation with neural networks. However, such methods can hardly be employed for cross-domain aspect category detection task because different domains always have different feature spaces, data distributions, and aspect spaces.

Heterogeneous transfer learning \cite{weiss2016survey} was proposed for the non-equivalent of feature spaces or label spaces. However, according to the transfer learning survey \cite{day2017survey}, few existing methods addressed the issue of differing label spaces. Furthermore, the existing methods, which can directly address the issue of differing label spaces, usually had additional restrictions. For instance, \cite{feuz2015transfer} required the construction of meta-features, and \cite{moon2016proactive} required the output label to be a pre-trained word embedding. Therefore, the existing heterogeneous transfer learning methods cannot be directly applied in the proposed cross-domain aspect category transfer and detection problem.

\textbf{Graph embedding}. Graph embedding algorithms, namely network representation learning models, aim to learn the low dimensional feature representations of nodes in networks. Although the techniques utilized in the models are different, most existing graph embedding models focus more on local graph structure representation, e.g., DeepWalk \cite{perozzi2014deepwalk} and Node2vec \cite{grover2016node2vec} considered a fixed-size context window of random walk generated node sequences; LINE \cite{tang2015line} modeled first- and second-order graph neighbourhood; GCN \cite{kipf2017semi} used convolutional operation to capture the local adjacency information. Moreover, the above algorithms were all designed for homogeneous graphs that might experience problems when applied to a heterogeneous graph. Metapath2vec++ \cite{dong2017metapath2vec} designed a global random walk pattern in heterogeneous graph to enhance the representation performance. But this method relies on human defined rules, which could be time-consuming, incomplete, and biased.

Unlike prior studies, we utilized e-commerce user behavior information to bridge the gap between the source and target domains. The proposed THGRL method enables the graphical aspect information transfer, which can not only project different domains' feature spaces into a common one but also allow data distributions and output spaces stay differently. Meanwhile, THGRL is fully automatic without handcrafting feature usage. To the best of our knowledge, few existing studies have investigated the user behavior information with graphical approach for cross-domain aspect category detection problem, and the proposed heterogeneous graph mining algorithm is innovative.

%
\section{Conclusion}

In this paper, we propose a traceable heterogeneous graph representation learning model (THGRL) for cross-domain aspect category transfer and detection. Unlike most of the prior studies, which only employ text information, THGRL leverages user behavior information, which offers important potential to address the cold-start problem. The proposed model can project heterogeneous objects (aspect, review, word, product, customer, and seller) from different domains into a joint embedding space. An innovative latent variable ``Walker Tracer'' is introduced to characterize the global semantic/aspect dependencies and capture the informative vertexes on the random walk paths.

The performance of the proposed method is comprehensively evaluated in three real-world datasets with four challenging tasks. The experimental results show that the proposed model significantly outperforms a series of state-of-the-art methods. Meanwhile, the case study empirically proves that the proposed model can successfully capture the global semantic/aspect dependencies (coherency pattern) in the heterogeneous graph, which is essential for cross-domain aspect knowledge transfer to overcome the problems of different data distributions and diverse output spaces.

The proposed THGRL algorithm is an unsupervised graph embedding model for heterogeneous graph mining. Theoretically, other tasks can adopt this algorithm as long as the dataset can be represented in a heterogeneous graph form. 

In the future, we will explore the proposed method on other heterogeneous graph based tasks, e.g., product or social recommendations. Meanwhile, we will investigate a more sophisticated method to improve the heterogeneous graph representation learning performance, such as enabling personalized heterogeneous graph navigation for random walk optimization. Furthermore, we will conduct in-depth studies of the user behavioral information, such as the impact of different entities (seller, customer, etc.) or different relations ($C\overset{wri}{\rightarrow} R$, $S\overset{get}{\rightarrow} R$, etc.) on cross-domain aspect transfer problem.

\begin{acks}
This work is supported by the National Natural Science Foundation of China (61876003, 81971691), the China Department of Science and Technology Key Grant (2018YFC1704206), and Fundamental Research Funds for the Central Universities
(18lgpy62).
\end{acks}

\bibliographystyle{ACM-Reference-Format}
\bibliography{refs}

\end{document}